\pdfoutput=1

\documentclass[11pt]{article}

\usepackage[final]{acl}
\usepackage{times}
\usepackage{latexsym}
\usepackage{algorithm}
\usepackage{algpseudocode}
\usepackage{amsmath}
\usepackage{multirow} 
\usepackage{subcaption}
\usepackage{booktabs}      
\usepackage{caption}       
\usepackage{graphicx} 
\usepackage{xcolor}

\usepackage[T1]{fontenc}

\usepackage[utf8]{inputenc}

\usepackage{microtype}

\usepackage{inconsolata}

\usepackage{graphicx}
\usepackage{subcaption} 
\usepackage{hyperref}

\title{Simulating Student-Tutor Interactions with LLMs \\ for Multilingual Math Feedback}

\author{
    \textbf{Junior Cedric Tonga}\textsuperscript{1}\thanks{
        Correspondence to: \texttt{junior.tonga@mbzuai.ac.ae}
    } \quad
    \textbf{KV Aditya Srivatsa}\textsuperscript{2}\thanks{
        Work done while at MBZUAI.
    } \quad
    \textbf{Kaushal Kumar Maurya}\textsuperscript{3}\footnotemark[2] \\
    \textbf{Fajri Koto}\textsuperscript{1} \quad
    \textbf{Ekaterina Kochmar}\textsuperscript{4}\footnotemark[2] \\[0.5em]
    \textsuperscript{1}Mohamed bin Zayed University of Artificial Intelligence \\
    \textsuperscript{2}ETH Z\"urich
    \textsuperscript{3}IIT Delhi Abu Dhabi
    \textsuperscript{4}University of Victoria
}

\begin{document}
\maketitle

\begin{abstract}


Large language models (LLMs) have demonstrated the ability to generate formative feedback and instructional hints in English, making them increasingly relevant for AI-assisted education. However, their ability to provide effective instructional support across different languages, especially for mathematically grounded reasoning tasks, remains largely unexamined. In this work, we present the \textit{first} large-scale simulation of multilingual student-tutor interactions using LLMs. A stronger model plays the role of the tutor, generating feedback in the form of hints, while a weaker model simulates the student. We explore 352 experimental settings across 11 typologically diverse languages, four state-of-the-art LLMs, and multiple prompting strategies to assess whether language-specific feedback leads to measurable learning gains.\footnote{\href{https://github.com/JuniorTonga/Simulating-Student-Tutor-Interactions-with-LLMs-for-Multilingual-Math-Feedback}{Code on GitHub}} Our study examines how student input language, tutor feedback language, model choice, and language resource level jointly influence performance. 
Results show that English hints are generally preferred, even by multilingual student LLMs, with few exceptions that lead to improved student learning outcomes. These findings offer practical insights for developing multilingual, LLM-based educational tools that are both effective and inclusive.


\end{abstract}

\section{Introduction}
\label{sec:intro}


Large language models (LLMs) have demonstrated strong reasoning abilities in solving mathematical problems, particularly when prompted in English \cite{kojima2022large, guo2025deepseek, bandyopadhyay2025thinking}. A common benchmark for evaluating such capabilities is {\tt GSM8K} \cite{Cobbe2021TrainingVT}, which consists of grade-school-level math word problems. Its multilingual counterpart, {\tt MGSM8K} \cite{shi2022mgsm}, extends this evaluation to a typologically diverse set of languages. Both English-centric and multilingual LLMs perform substantially worse on {\tt MGSM8K} than on its English-only {\tt GSM8K} benchmark \cite{shi2022mgsm, ko2025ust}, highlighting a gap in LLMs' cross-linguistic reasoning abilities. This discrepancy raises questions about the use of LLMs as instructional agents beyond English. Recent work has explored their role as proxy tutors, generating formative feedback and pedagogical hints to support weaker student models or human learners \cite{wang2024tutor, meyer2024using}. One widely studied form of support is hinting: a concise prompt aimed at guiding problem-solving without directly providing the answer. While such interventions have been shown to improve learning outcomes in English~\cite{kochmar2022automated}, their impact in multilingual settings remains largely unexamined. Moreover, these explorations are conducted in isolation, where either the tutor generates a hint or the student merely consumes it.

In this work, we take a step further by {\em simulating student-tutor interactions} entirely using LLMs: a stronger model generates hints as a tutor, while a weaker model attempts to solve the problem as a student. This simulation setup allows us to explore the effects of language, hint quality, and prompting strategy in a scalable and reproducible way. Moreover, such LLM-to-LLM simulations can serve as a valuable proxy for real-world educational scenarios, offering insights into how multilingual feedback might impact learning before deploying these systems with actual students. To the best of our knowledge, this is \textbf{the first work to simulate multilingual student-tutor interactions between LLMs across a broad range of languages and settings}. Given that effective feedback depends on both linguistic and reasoning proficiency, this raises a central question: \textit{Does multilingual feedback from LLM tutors lead to measurable learning gains in student models?}

Our key contributions are as follows:
\begin{enumerate}
    \itemsep0em
    \item We simulate student-tutor interactions entirely using LLMs, modeling multilingual feedback across 11 languages, multiple prompting strategies, and four LLMs, yielding a large-scale experimental space of 352 settings.
    \item We investigate how language-specific feedback influences student performance, examining the interplay between student input language, tutor feedback language, model selection, and whether the language is high- or low-resource, within the domain of mathematically grounded reasoning tasks.
    \item We offer practical recommendations for designing LLM-based systems that support effective multilingual feedback and hint generation, highlighting considerations for both research and real-world educational deployment.
\end{enumerate}



\begin{figure*}[t]
    \centering
    \includegraphics[scale=0.45]{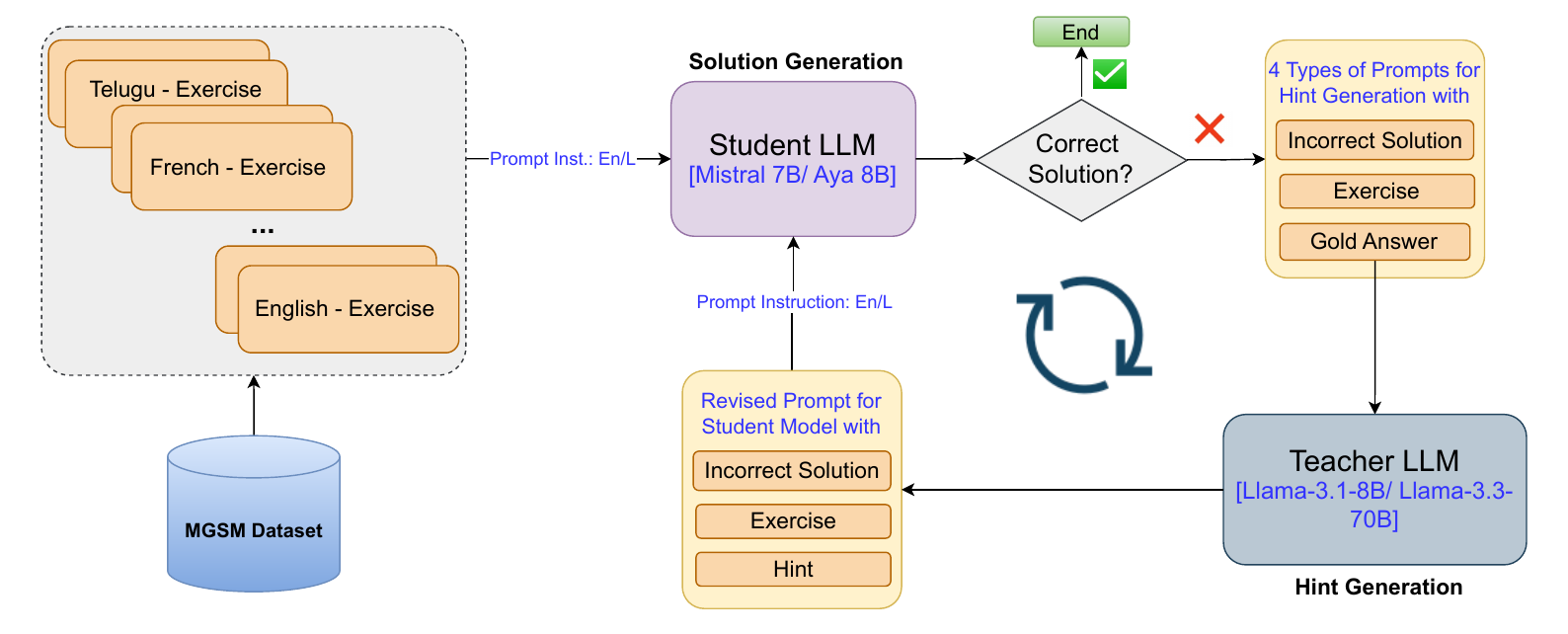}
    \vspace{-0.3cm}
   \caption{Overview of the student-tutor interaction flow.}
    \label{fig:flow_digm}
    \vspace{-0.6cm}
\end{figure*}


\section{Related Work}
\label{sec:related}

\paragraph{LLMs in Math Reasoning}
LLMs have demonstrated strong performance in mathematical reasoning tasks, particularly in English, with benchmarks like GSM8K \cite{Cobbe2021TrainingVT} driving much of this progress. 
Specialized models such as Minerva \cite{lewkowycz2022solving}, trained in scientific texts, achieve state-of-the-art results without relying on external tools. However, these advances are still largely focused on English. Recent benchmarks like MGSM8K \cite{shi2022mgsm} reveal that multilingual LLMs underperform significantly, especially in low-resource languages, due to limited language coverage and weaker alignment between linguistic and mathematical representations.


\paragraph{Automated Hint Generation}
Before neural language models, automatic hint generation was typically formulated as a Markov Decision Process (MDP), selecting optimal hints based on student states \citep{stamper2008hintfactory}. Subsequent work improved scalability by structuring large hint sets into solution paths, and \citet{paassen2017continuous} advanced this by modeling hint policies in continuous edit-distance spaces for better generalization. With LLMs, research has shifted toward direct hint generation: GPT-4 has been paired with a GPT-3.5 ``student-validator'' to filter poor hints \citep{phung2023gpt4hints}, and smaller models like LLaMA-3-8B were shown to approach GPT-4o performance when prompts target specific error types \citep{pmlr-v264-tonga25a}. Studies show ChatGPT-generated hints can yield learning gains comparable to human hints in mathematics \citep{pardos2024chatgpt}, though quality varies by domain and task complexity, and \citet{mcnichols2024can} note LLMs still struggle to generalize beyond training data to unseen student errors.


However, much of prior research has focused exclusively on English. This narrow scope limits the applicability of LLM-based tutoring in multilingual learning environments, where students often benefit more from feedback in their native language. Recent work such as MathOctopus \citep{chen-etal-2024-breaking} shows that multilingual tuning can significantly improve math reasoning across languages.
However, the generation of effective instructional hints in languages beyond English, especially for low-resource contexts, remains an open challenge.

\paragraph{Simulating student behavior with LLMs.}
Although persona simulation has long been studied, it has recently gained momentum in LLM-based multi-agent systems for modeling agents with distinct personalities and goal-driven behaviors \cite{tran2025multi}. In education, early approaches like SimStudent \citep{matsuda2013simstudent} aimed for faithful student simulation. Modern LLMs allow for richer, controllable simulations: \citet{benedetto2024simulate} adjust skill levels when generating exam answers, MathVC creates virtual classrooms with LLM “students” exhibiting diverse misconceptions \citep{yue2024mathvc}, and error-focused datasets like LEMMA help models learn to self-repair \citep{rao2025lemma}. Recent work, such as \citet{macina2023mathdial}, also explores multi-turn dialogues where LLMs act as students in conversations with human tutors. These developments provide ethically appropriate alternatives to real students. \textit{Our work builds on these efforts by treating LLMs as simulated students, aligned with established guidelines and best practices.}

\section{Methodology}
This section describes our modeling framework, system architecture, and prompting strategies to simulate student-tutor interactions with LLMs. The overall design is illustrated in Figure~\ref{fig:flow_digm} and consists of the following key components:

\paragraph{Solution Generation} 
The first stage of our pipeline involves generating a candidate student solution to a given multilingual exercise. Let $x \in \mathcal{X}$ represent an exercise in a given language $L$, and let $LLM_S$ denote the student language model. The model is prompted with $x$ to produce a candidate solution $\hat{y}_S = LLM_S(x)$. To mimic real-world learner behavior, we consider multilingual scenarios where the instructional prompt is given either in English or in the native language of $x$ (i.e., $L$). This design choice allows us to evaluate the language sensitivity of $LLM_S$ and its downstream performance. 
We experiment with two student models: the instruction-tuned {\tt Mistral-7B} and the multilingual {\tt Aya-8B}, chosen for their balance between model capacity and efficiency.

\vspace{-0.2cm}
\paragraph{Hint Generation} For exercises where the generated solution $\hat{y}_S$ is found to be incorrect (i.e., $\hat{y}_S \ne y^*$, where $y^*$ is the reference or gold solution), we employ a tutor model $LLM_T$ to generate pedagogically helpful hints. The motivation here is to simulate intelligent tutoring interventions that guide students toward the correct solution path without directly revealing the answer. The tutor model takes as input a triplet $\langle x, \hat{y}_S, y^* \rangle$ and produces a hint $h = LLM_T(x, \hat{y}_S, y^*)$ under one of four controlled prompting strategies (described in the next paragraph on prompting). 
GPT-4o \citep{Hurst2024GPT4oSC} was used to validate the correctness of both the initial and revised solutions by comparing them to the gold solution (see prompts in Appendix \ref{subsec:eval_prompt}).
We employ large {\tt LLaMA-3.3-70B} LLM as the main tutor but also experiment with the smaller {\tt LLaMA-3.1-8B} LLM.

\vspace{-0.1cm}
\paragraph{Prompting}
We consider two prompting setups for the $LLM_S$: (1)~\textit{\textbf{Multilingual prompting}}, where the student prompt is written in the same language $L$ as the exercise. To operationalize this, we translate the base prompts (provided in Figure~\ref{fig:prompt_init_sol} and Figure~\ref{fig:prompt_revise}) into the 11 languages of the MGSM benchmark (detailed discussion in Section \ref{sec:exp_setup}) using Google Translate API;\footnote{Translation performed using Google Translate API from \url{https://github.com/nidhaloff/deep-translator}.} (2)~\textit{\textbf{English-only prompting}}, where the student prompt remains in English, regardless of the exercise's language $L$. This serves as a control condition to isolate the impact of prompt language on downstream performance.

For hint generation, we explore four strategies by varying the \textbf{input prompt language} to the tutor model $LLM_T$ and the \textbf{output hint language}:

\begin{enumerate}
    \itemsep-0em
    \item \textbf{English-to-English (EN$\to$EN)}: The tutor model ($LLM_T$) is prompted in English and instructed to generate a hint in English, regardless of the language of the original exercise. This setup uses the prompt format illustrated in Figure~\ref{fig:prompt_generate_hint} of the Appendix.
\label{list:hint-strategies}
    \item \textbf{English-to-English with Translation (EN$\to$EN$\to$L)}: The hint generated in the EN$\to$EN setup is machine-translated with Google Translate into the exercise’s native language $L$. This configuration controls for content while varying the delivery language, enabling analysis of whether presenting hints in the student's native language improves comprehension and learning outcomes.

    \item \textbf{Native-to-Native (L$\to$L)}: The tutor model is prompted in the native language $L$ of the exercise, using the translated version of the hint prompt provided in Appendix (see Figure~\ref{fig:prompt_generate_hint}), and is instructed to generate the hint in the same language $L$. This aims to explore the impact of native language interaction and hints on the tutor model.  

    \item \textbf{English-to-Native (EN$\to$L)}: The tutor model is prompted in English, following the format shown in Figure~\ref{fig:prompt_generate_hint} of the Appendix, but is explicitly instructed to generate the hint in the target language $L$. This explores how instruction in English and hinting in the native language affects the tutoring outcomes. 
\end{enumerate}

\paragraph{Student-Tutor Interaction Flow}
The full pipeline of student-tutor interaction is summarized in Algorithm~\ref{alg:abstract_flow}. It outlines the interaction between the student and tutor models, including the hint-guided revision loop. Due to computational constraints, we primarily experimented with a single hint iteration ($N = 1$). However, we also include a small-scale analysis for $N>1$ (see Section \ref{sec:fur_ana}), which shows that the key observations made with $N = 1$ largely hold across higher values of $N$ as well.

\begin{algorithm}[ht]
\small 
\caption{Student-Tutor Interaction Flow}
\label{alg:abstract_flow}
\begin{algorithmic}[1]
\Require Exercise $x$ in language $L$, reference solution $y^*$, maximum hint attempts $N$
\State Choose student prompting mode ($P_{S}$): \textit{Multilingual} or \textit{English-only}
\State Generate solution $\hat{y}_S \gets LLM_{S}(P_S(x))$
\If{$\hat{y}_S = y^*$}
    \State \Return Correct solution
\EndIf

 \Comment{\textit{Hint-Guided Revision Loop}}
\For{$i = 1$ to $N$}
    \State Choose hint generation prompt strategy $P_T$
    \State Generate hint $h \gets LLM_{T}(P_T(x, \hat{y}_S, y^*))$
    \State Provide hint $h$ to $LLM_S$ and generate revised solution $\hat{y}_S \gets LLM_{S}(P_S(x, \hat{y}_S, h))$
    \If{$\hat{y}_S = y^*$}
        \State \Return Correct solution
    \EndIf
\EndFor

\State \Return Final student solution $\hat{y}_S$ after $N$ attempts
\end{algorithmic}
\end{algorithm}
\vspace{-0.5cm}

\section{Experimental Setup}
\label{sec:exp_setup}
\vspace{-0.2cm}


\paragraph{Dataset and Languages}
We use the Multilingual Grade School Math ({\tt MGSM}) dataset \citep{shi2022mgsm}, a multilingual extension of {\tt GSM8K} \citep{Cobbe2021TrainingVT}, consisting of 250 grade-school-level math problems translated from English into 11 typologically and geographically diverse languages (English (en), Bengali (bn), Chinese (zh), French (fr), German (de), Japanese (ja), Russian (ru), Spanish (es), Swahili (sw), Telugu (te), and Thai (th)). These examples are representative of the full {\tt GSM8K} dataset (see Appendix~\ref{sec:app-data-representation}). Following prior work \cite{shi2022mgsm}, we group the languages into High-Resource Languages (HRLs: en, zh, fr, de, ja, ru, es) and Low-Resource Languages (LRLs: bn, th, te, sw).
\paragraph{Models}
We use four LLMs: two tutor models, \texttt{LLaMA-3.3-70B} and \texttt{LLaMA-3.1-8B}, selected for their multilingual strength, and two student models, \texttt{Mistral-7B} (monolingual) and \texttt{Aya-8B} (multilingual). These cover a range of model sizes and capabilities. Appendix Section~\ref{sec:compare_anlysis} provides solution accuracy scores.

\paragraph{Evaluation Metric}
We use student gain $G$ as the main evaluation metric, defined as the relative improvement in a student model's accuracy after receiving hints. Let \( A_{\text{before}} \) and \( A_{\text{after}} \) denote accuracy before and after hinting (over $N$ iterations). The Student Gain \( G \) is:
\vspace{-0.5cm}
\begin{center}
\small
\begin{minipage}{0.45\linewidth}
\[
G = \frac{A_{\text{after}} - A_{\text{before}}}{A_{\text{before}}} \times 100
\]
\end{minipage}
\hfill
\begin{minipage}{0.45\linewidth}
\[
\bar{G}_{\text{category}} = \frac{1}{L} \sum_{i=1}^L G_i
\]
\end{minipage}
\end{center}
This percentage-based metric accounts for the model’s initial performance \(A_{\text{before}} \) and allows for meaningful comparisons ~\cite{tornqvist1985should}. We average the gain G across all languages within each category (i.e, HRLs and LRLs) to obtain $\bar{G}_{\text{category}}$,
with \( L \) being the number of languages in the category, and \( G_i \) is the gain for language \( i \).




\paragraph{Experimental Space} 
Our setup spans 11 languages, 2 student models, 2 prompt types, 2 tutor models, and 4 hint strategies—yielding \textbf{352 unique configurations }(see Table~\ref{tab:experiment-space} in Appendix). See Appendix Section~\ref{sec:impl_details} for implementation details.

\section{Results}


In this section, we discuss the impact of different student models and prompts, tutor models and prompts, and variations of student gains across HRLs and LRLs. We present key observations relevant to our core research question on multilingual feedback in mathematical reasoning tasks.



\begin{table}[t]
\centering
\resizebox{\columnwidth}{!}{%
\begin{tabular}{llcccc}
\toprule
& & \textbf{En→En} & \textbf{En→En→L} & \textbf{En→L} & \textbf{L→L} \\
\midrule
\multicolumn{6}{l}{\textit{\textsc{Multilingual Setup:} Student input prompt in native language}} \\
\midrule
\multirow{2}{*}{HRLs} & {\tt Mistral-7B} & 11.30 & 10.90 & 9.50 & 11.00 \\
                      & {\tt Aya-8B}     & 4.00  & 3.60  & 3.60 & 3.90  \\
\midrule
\multirow{2}{*}{LRLs} & {\tt Mistral-7B} & 12.10 & 6.60  & 5.60 & 14.90 \\
                      & {\tt Aya-8B}     & 27.60 & 16.90 & 17.40 & 20.50 \\
\midrule
\textbf{Avg. HRLs} & & \textcolor{blue}{7.65} & 7.25 & 6.55 & \textcolor{green!70!black}{7.45} \\
\textbf{Avg. LRLs} & & \textcolor{blue}{19.85} & 11.75 & 11.50 & \textcolor{green!70!black}{17.70} \\
\midrule
\textbf{Avg. Overall} & & 13.75 & 9.50 & 9.03 & 12.57 \\
\midrule
$\Delta$ HRLs & {\tt Mistral-7B}$-${\tt Aya-8B} & 7.30 & 7.30 & 5.90 & 7.10 \\
$\Delta$ LRLs & {\tt Mistral-7B}$-${\tt Aya-8B} & -15.50 & -10.30 & -11.80 & -5.60 \\
\midrule
\multicolumn{6}{l}{\textit{\textsc{English-Only Setup:} Student input prompt in English}} \\
\midrule
\multirow{2}{*}{HRLs} & {\tt Mistral-7B} & 8.70 & 11.00 & 12.10 & 10.60 \\
                      & {\tt Aya-8B}     & 5.90 & 5.50  & 4.70  & 4.70  \\
\midrule
\multirow{2}{*}{LRLs} & {\tt Mistral-7B} & 46.00 & 7.00  & 43.10 & 8.80  \\
                      & {\tt Aya-8B}     & 35.00 & 15.10 & 13.40 & 15.70 \\
\midrule
\textbf{Avg. HRLs} & & 7.30 & \textcolor{green!70!black}{8.25} & \textcolor{blue}{8.40} & 7.65 \\
\textbf{Avg. LRLs} & & \textcolor{blue}{40.50} & 11.05 & \textcolor{green!70!black}{28.25} & 12.25 \\
\midrule
\textbf{Avg. Overall} & & 23.90 & 9.64 & 18.33 & 9.95 \\
\midrule
$\Delta$ HRLs & {\tt Mistral-7B}$-${\tt Aya-8B} & 2.80 & 5.50 & 7.40 & 5.90 \\
$\Delta$ LRLs & {\tt Mistral-7B}$-${\tt Aya-8B} & 11.00 & -8.10 & 29.70 & -6.90 \\
\bottomrule
\end{tabular}%
}
\caption{Mean student gains (\%) across two experimental setups using {\tt LLaMA-3.3-70B} as the tutor model. The \textcolor{blue}{best} average values and \textcolor{green!70!black}{second-best} values are highlighted. $\Delta$ rows indicate performance differences between {\tt Mistral-7B} and {\tt Aya-8B} within each language resource category (HRLs and LRLs).}
\label{tab:strategie_comparison}
\vspace{-0.5cm}
\end{table}

\paragraph{Does the size of the tutor model impact student gains?}

Based on Appendix Figure \ref{sec:det_result}, we observe that {\tt LLaMA-3.3-70B} consistently yields greater gains across both student models and hint generation prompts, outperforming {\tt LLaMA-3.1-8B} for HRLs and LRLs. Specifically, {\tt LLaMA-3.3-70B} achieves higher median gains -- for instance, 8.6\% in the \textit{multilingual} prompt setup and 10\% in the \textit{English-only} with {\tt Mistral} -- compared to 7.7\% and 8\% for {\tt LLaMA-3.1-8B}. This effect is especially pronounced for LRLs, with median improvements reaching up to 31\% ({\tt Aya-8B}, \textit{multilingual}) and 38\% ({\tt Mistral-7B}, \textit{English-only} setup). While {\tt LLaMA-3.1-8B} also shows strong improvements for LRLs, its performance exhibits higher variability than {\tt LLaMA-3.3-70B}, as evidenced by the presence of outliers and larger interquartile ranges in the boxplots, indicating a less equitable distribution for LRLs. \textit{These results suggest that larger tutor models are more effective at generating helpful hints, and that model size plays a key role in mitigating the challenges of low-resource settings.} This is expected as larger models are more capable overall. To simplify the interpretation of our findings, we focus the remainder of our discussion on the large tutor model only.

\paragraph{Does the multilingual student model prompting lead to higher student gains compared to English-only prompting?}
The \textit{Avg.}~labeled rows from Table~\ref{tab:strategie_comparison} (distilled from Figure~\ref{fig:box_plot_improvement}) show that \textit{English-only} prompting consistently outperforms \textit{multilingual} prompting across all hint types, except for the L→L setting. This indicates that delivering instructions in English is generally beneficial regardless of hint language. However, when instructions are provided in a native language, the corresponding hints should also be in the native language, especially for LRLs, where this effect is more pronounced. For HRLs, the average gain is comparable across all hint prompts for both setups. Finally, these observations are largely invariant to the type of student LLM, whether monolingual or multilingual. \textit{Overall, multilingual instruction does not necessarily lead to higher student gains unless paired carefully with the hint language.}

\paragraph{Which student model leads to greater gains -- multilingual or English-centric?}
The rows labeled with $\Delta$ in Table~\ref{tab:strategie_comparison} provide clarity on this question. A negative $\Delta$ indicates better performance by the multilingual {\tt Aya-8B} model, while a positive value favors the monolingual {\tt Mistral-7B}. For LRLs, {\tt Aya-8B} tends to be more effective, likely due to stronger language representation. In contrast, {\tt Mistral-7B} generally performs better on HRLs. Interestingly, the difference in student gains is more pronounced on the average in the \textit{multilingual} setting compared to the \textit{English-only} setting, as indicated by higher absolute values of $\Delta$ scores. \textit{Overall, as expected, multilingual LLMs tend to perform better for LRLs, while monolingual LLMs are better suited for HRLs.}

\paragraph{Which hint generation prompt strategy performs best with {\tt \textbf{LLaMA-3.3-70B}}?}
Figure~\ref{fig:box_plot_improvement} and Table~\ref{tab:strategie_comparison} present the comparative performance of various hint generation strategies under both \textit{multilingual} and \textit{English-only} settings. The EN→EN strategy yields the highest average improvement across both setups, outperforming other hint prompting strategies. This indicates that models are highly responsive to English prompts. This trend generally holds across both HRLs and LRLs, as well as across different student LLM types. An exception is observed in the \textit{English-only} setting for HRLs when using {\tt Mistral-7B} as the student LLM, where EN→EN is less effective, possibly due to limited representation of HRLs in {\tt Mistral-7B}. \textit{Overall, tailoring hint strategies to model- and language-specific characteristics is essential for maximizing student gains, though EN→EN remains the most preferred strategy.}

\paragraph{Does the language of the hint (English vs. the native) influence student gains?}

Using the larger tutor model, {\tt LLaMA-3.3-70B}, we find that hint language substantially impacts student model gains as shown in Table \ref{tab:strategie_comparison}. In the \textit{multilingual} prompt setup, English hints (generated using the EN→EN tutor prompt) consistently yield the highest median improvements across both student models. Further, for LRLs, {\tt Aya-8B} achieves its best and most consistent improvement of 31.12\%, while {\tt Mistral-7B} reaches its highest median improvement of 7.6\%. In the \textit{English-only} setup, the trend holds: {\tt Mistral-7B} and {\tt Aya-8B} obtain their best results for LRLs with English hints, reaching 38.19\% and 16.24\%, respectively. For HRLs, {\tt Aya-8B} performs slightly better with translated hints using the EN→EN→L prompt, whereas {\tt Mistral-7B} benefits more from hints provided directly in the target language (L→L). In contrast, when using the smaller tutor model, {\tt LLaMA-3.1-8B}, no consistent patterns emerge regarding hint language effectiveness. \textit{Overall, English-language hints tend to be more effective when generated by larger tutor models. However, native-language hints can be competitive or even superior in specific cases.}

\paragraph{Student gains across languages.}
The Avg. row in Table~\ref{tab:strategie_comparison} highlights that student models achieve higher gains on LRLs than HRLs in both \textit{multilingual} and \textit{English-only} setups. Multilingual LLMs perform better on LRLs, while monolingual LLMs are more effective for HRLs, consistent with earlier findings. Additionally, among the four hint prompting strategies, the EN→EN prompt yields the highest overall gains across all languages. Further, Table \ref{tab:student_gain_mistral} (Appendix) shows that {\tt Mistral-7B} struggles with Telugu and Bengali in the \textit{multilingual} setup but achieves better gains in these languages in the \textit{English-only} setup (EN→EN, EN→L). In contrast, {\tt Aya-8b} demonstrates more consistent and higher gains across LRLs (Telugu, Swahili, Bengali, and Thai) in both setups (see Table \ref{tab:student_gain_aya}, Appendix).

\paragraph{Final Takeaways}
Based on the results and discussion, we summarize our key findings regarding multilingual student–tutor interactions:

\begin{enumerate}
\itemsep-0em
\item \textit{Student Prompt:} {\tt English-only} prompts generally perform well; however, when either the hint generation prompt or the hint is in the native language, multilingual prompting may be preferable.
\item \textit{Student Model:} Monolingual models perform better for HRLs, while multilingual models are more effective for LRLs.
\item \textit{Hint Generation Prompt:} EN→EN remains the most preferred strategy, with a few exceptions.
\item \textit{Tutor Model:} Larger models such as {\tt LLaMA-3.3-70B} are generally more effective and should be preferred.
\item \textit{Hint Language:} English hints are generally preferred; however, for HRLs and monolingual student models, native-language hints can be more effective.
\end{enumerate}

\section{Further Analyses}

\begin{table}[t]
\centering
\resizebox{\columnwidth}{!}{%
\begin{tabular}{ll|cc|cc|cc|cc}
\hline
& & \multicolumn{2}{c|}{\textbf{En→En}} & \multicolumn{2}{c|}{\textbf{En→En→L}} & \multicolumn{2}{c|}{\textbf{En→L}} & \multicolumn{2}{c}{\textbf{L→L}} \\
\hline
& \textbf{Models} & \textbf{L-3.1} & \textbf{L-3.3} & \textbf{L-3.1} & \textbf{L-3.3} & \textbf{L-3.1} & \textbf{L-3.3} & \textbf{L-3.1} & \textbf{L-3.3} \\
\hline
\multirow{2}{*}{\textbf{HRLs}} & \texttt{\textbf{Mistral-7B}} & 0.51 & 0.80 & 0.51 & 0.80 & 0.51 & 0.74 & \textbf{1.43} & 0.63 \\
& {\tt \textbf{Aya-8B}} & 0.46 & 0.51 & 0.46 & 0.51 & 0.17 & 0.29 & \textbf{0.57} & 0.06 \\
\hline
\multirow{2}{*}{\textbf{LRLs}} & \texttt{\textbf{Mistral-7B}} & 0.70 & 1.90 & 0.70 & 1.90 & 0.50 & 1.60 & \textbf{4.70} & 2.20 \\
& {\tt \textbf{Aya-8B}} & 1.10 & 1.80 & 1.10 & 1.80 & 1.20 & 1.30 & \textbf{8.30} & 1.60 \\
\hline
\end{tabular}%
}
\caption{\small Answer leakage proportions (\%) across {\tt LLaMa-3.1-8B} (as L-3.1) and {\tt Llama-3.3-70B} (as L-3.3).}
\label{tab:answer_leakage}

\end{table}

\paragraph{Gold Answer Leakage.}
Despite explicit instructions to the tutor model to avoid revealing the final answer while generating hints, gold answer leakage may still occur due to LLM hallucinations, potentially compromising our findings. To assess this risk, we conduct a \textit{Gold Answer Leakage} test—i.e., \textit{checking whether the final answer appears verbatim within the generated hint.} This is a challenging task, as it requires precise extraction of the gold answer from free-form hints; we adopt a regex-based approach for detection. During hint generation, we flagged any hints that included the gold answer as a \textit{stand-alone number} (i.e., not embedded in a longer number or decimal).  
Detection used the regex,\footnote{Regex pattern = \texttt{r'(?<!\.)}' + \texttt{r'\textbackslash b}' + \texttt{re.escape(answer\_str)} + \texttt{r'\textbackslash b}' + \texttt{r'(?!\.)'}} which matches the exact integer or decimal token when it is delimited by non-digit boundaries and not attached to a decimal point. Table~\ref{tab:answer_leakage} reports the proportion of such hints among the total samples for HRLs (250×7 = 1750) and LRLs (250×4 = 1000), across the four hint generation strategies and student models. The highest observed answer leakage was approximately 8\% for {\tt LLaMA-3.3-8B} and around 2\% for {\tt LLaMA-3.3-70B}. Since our primary tutor model is {\tt LLaMA-3.3-70B}, this level of leakage is unlikely to significantly impact our findings, especially considering that the regex-based extractors tend to produce some false positives. Notably, the leakage rate is higher for low-resource languages, suggesting greater difficulty in handling those languages.


We further investigated whether the hints that helped students revise their initial answers tended to contain the gold answer. Focusing on the two best-performing strategies in the \textit{multilingual} setup—EN→EN and L→L—we computed the leakage ratio, defined as the number of helpful hints that included the gold answer divided by the total number of helpful hints. These results are presented in Figure~\ref{fig:leakage}. The findings suggest that helpful hints rarely reveal the gold answer, with leakage ratios close to zero for most languages. Notable exceptions include Thai and Swahili LRLs, where over 2\% of helpful hints contained the gold answer.

\begin{table}[t]
\centering
\resizebox{\columnwidth}{!}{%
\begin{tabular}{ll|cc|cc|cc|cc}
\hline
& & \multicolumn{2}{c|}{\textbf{En→En}} & \multicolumn{2}{c|}{\textbf{En→En→L}} & \multicolumn{2}{c|}{\textbf{En→L}} & \multicolumn{2}{c}{\textbf{L→L}} \\
\hline
& \textbf{Models} & {\tt \textbf{Aya-8B}} & {\tt \textbf{Mistral-7B}} & {\tt \textbf{Aya-8B}} & {\tt \textbf{Mistral-7B}} & {\tt \textbf{Aya-8B}} & {\tt \textbf{Mistral-7B}} & {\tt \textbf{Aya-8B}} & {\tt \textbf{Mistral-7B}} \\
\hline
\multirow{2}{*}{\textbf{HRLs}} &\textbf{ L-3.1} & 99.99 & 99.88 & 99.88 & 99.88 & 99.31 & 97.62 & 99.87 & 99.92 \\
& \textbf{L-3.3} & 100.00 & 99.77 & 99.94 & 99.88 & 99.94 & 99.71 & 99.94 & 99.94 \\
\hline
\multirow{2}{*}{\textbf{LRLs}} &\textbf{ L-3.1} & 99.64 & 98.90 & 97.92 & \textbf{96.90} & 98.40 & 97.85 & 98.15 & 98.80 \\
& \textbf{L-3.3} & 100.00 & 99.80 & 98.95 & 99.17 & 99.40 & 99.37 & 99.47 & 99.40 \\
\hline
\end{tabular}%
}
\caption{\small Mean language identification accuracy for hints with  {\tt LLaMA-3.3-70B} (as L.3.3) and {\tt LLaMA-3.1-8B} (as L.3.1). The lowest number is bold.}
\label{tab:performance_scores}
\vspace{-0.3cm} 
\end{table}

\begin{table}[t]
\centering
\resizebox{0.9\columnwidth}{!}{%
\begin{tabular}{llcccc}
\hline
& & \textbf{En→En }& \textbf{En→En→L} & \textbf{En→L} & \textbf{L→L }\\
\hline
\multicolumn{2}{l}{\textbf{Initial solution}} & & & & \\
\hline
\multirow{2}{*}{\textbf{HRLs}} & {\tt Mistral-7b} & 99.8 & 99.8 & 99.8 & 99.8 \\
& Aya & 100.0 & 100.0 & 100.0 & 100.0 \\
\hline
\multirow{2}{*}{\textbf{LRLs}} & {\tt Mistral-7b} & 96.8 & 96.8 & \textbf{96.7} & 96.8 \\
& Aya & 98.6 & 98.6 & 98.6 & 98.6 \\
\hline
\multicolumn{2}{l}{\textbf{Revised solution}} & & & & \\
\hline
\multirow{2}{*}{\textbf{HRLs}} & {\tt Mistral-7b} & 93.1 & 93.7 & 94.9 & 94.3 \\
& Aya  & 92.0 & 92.9 & 93.4 & 93.0 \\
\hline
\multirow{2}{*}{\textbf{LRLs}} & {\tt Mistral-7b} & \textbf{90.6} & 92.1 & 92.3 & 91.8 \\
& Aya & 99.4 & 99.1 & 98.7 & 98.9 \\
\hline
\end{tabular}%
}
\caption{Mean language identification accuracy (in \%) of student models’ initial and revised solutions. Lowest value is bold for both initial and revised solutions.}
\vspace{-0.4cm} 
\label{tab:solution_comparison}
\end{table}

\paragraph{Language Consistency of Hints and Student Outputs.}
\textit{Are the initial solution, generated hints, and revised solution in the intended language?} To verify this, we conducted sanity checks using a FastText-based language identification (LID) method \cite{bojanowski-etal-2017-enriching} with the pre-trained \texttt{lid.176} model. Table~\ref{tab:performance_scores} reports the mean LID accuracy of hints for HRLs and LRLs across hint generation strategies and tutor models. With a minimum LID accuracy of $\sim$97\%, most hints are in the intended language with minor code-mixing.
Table~\ref{tab:solution_comparison} presents LID accuracy for initial and revised student solutions. Initial solutions are mostly in the expected language, with a minimum LID accuracy of $\sim$97\%. However, revised solutions show a drop in accuracy to $\sim$91\%, indicating increased code-mixing or language switching. Manual inspection reveals that both tutor models struggle to preserve language consistency in revised solutions, particularly for LRLs—most notably {\tt Mistral-7B}, likely due to its English-centric bias. This degradation may propagate from minor code-mixing in the hints themselves (as previously observed). Interestingly, the EN→L strategy yields the highest LID accuracy in revised outputs, suggesting it is more effective in maintaining language fidelity.

\paragraph{Translation Quality.}
To evaluate the quality of translated hints in the En→En→L setting—where English hints are translated into target languages using Google Translate—we perform back-translation to English and compute BLEU scores \cite{papineni-etal-2002-bleu} over all samples for each language. As shown in Table~\ref{tab:bleu_scores}, translation quality is generally high across languages, with the exception of Telugu, which shows only moderate quality. This further validates the findings with the En→En→L prompt. 

\begin{figure}
    \centering
    \includegraphics[width=\linewidth]{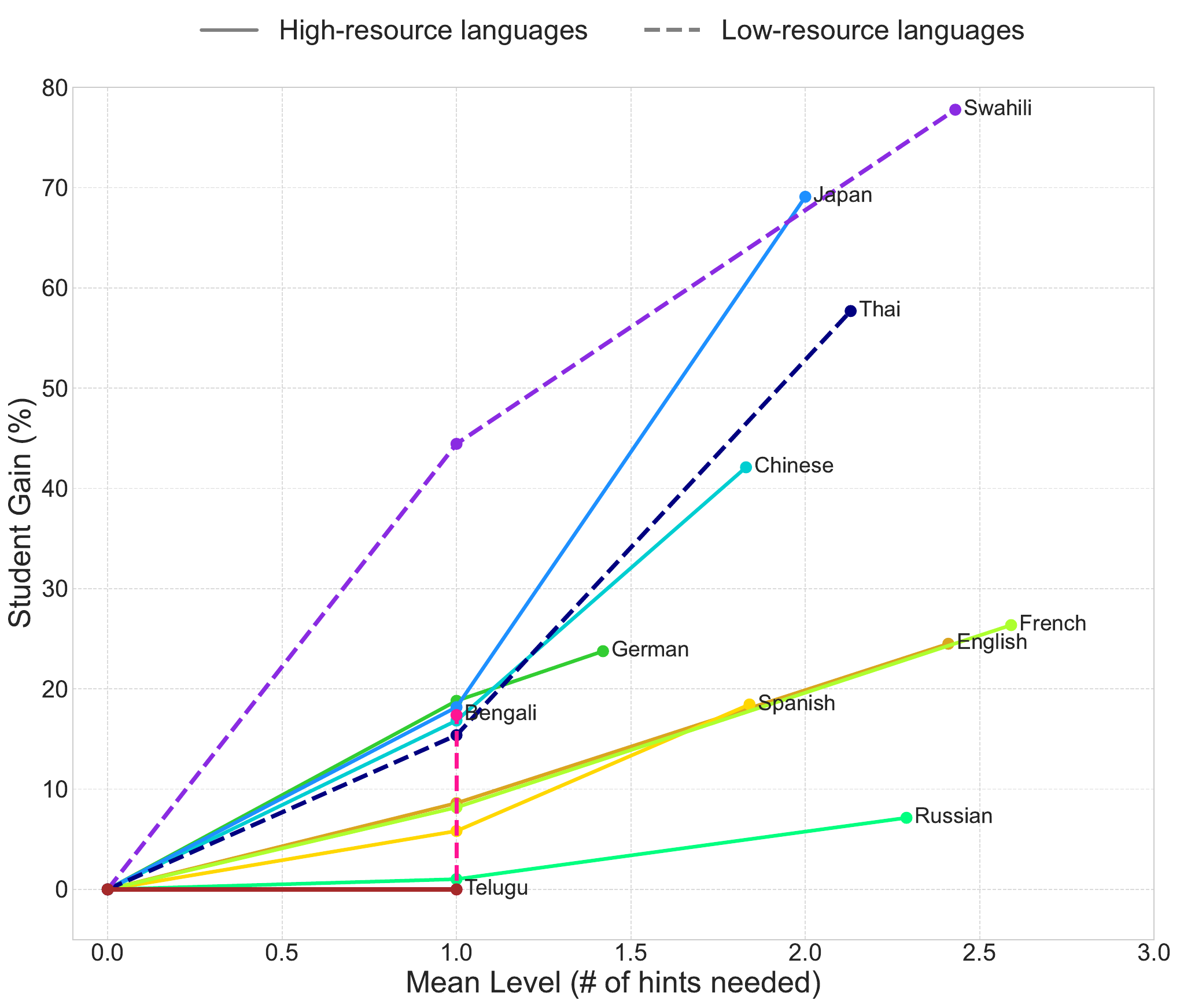}
    \caption{Student Gain scores as a function of the mean number of hints required per language.}
    \label{fig:5_hints}
\end{figure}

\begin{table}[t]
\centering
\resizebox{\columnwidth}{!}{%
\begin{tabular}{ll|cc|cc|cc|cc}
\hline
& & \multicolumn{2}{c|}{\textbf{En→En}} & \multicolumn{2}{c|}{\textbf{En→En→L}} & \multicolumn{2}{c|}{\textbf{En→L}} & \multicolumn{2}{c}{\textbf{L→L}} \\
\hline
& \textbf{Models} & {\tt \textbf{Aya-8B}} & {\tt \textbf{Mistral-7B}} & {\tt \textbf{Aya-8B}} & {\tt \textbf{Mistral-7B}} & {\tt \textbf{Aya-8B}} & {\tt \textbf{Mistral-7B}} & {\tt \textbf{Aya-8B}} & {\tt \textbf{Mistral-7B}} \\
\hline
\multirow{2}{*}{\textbf{HRLs}} &\textbf{ L-3.1} & 99.99 & 99.88 & 99.88 & 99.88 & 99.31 & 97.62 & 99.87 & 99.92 \\
& \textbf{L-3.3} & 100.00 & 99.77 & 99.94 & 99.88 & 99.94 & 99.71 & 99.94 & 99.94 \\
\hline
\multirow{2}{*}{\textbf{LRLs}} &\textbf{ L-3.1} & 99.64 & 98.90 & 97.92 & \textbf{96.90} & 98.40 & 97.85 & 98.15 & 98.80 \\
& \textbf{L-3.3} & 100.00 & 99.80 & 98.95 & 99.17 & 99.40 & 99.37 & 99.47 & 99.40 \\
\hline
\end{tabular}%
}
\caption{\small Mean language identification accuracy for hints with  {\tt LLaMA-3.3-70B} (as L.3.3) and {\tt LLaMA-3.1-8B} (as L.3.1). The lowest number is bold.}
\label{tab:performance_scores}
\vspace{-0.3cm} 
\end{table}

\begin{table}[t]
\centering
\resizebox{0.9\columnwidth}{!}{%
\begin{tabular}{llcccc}
\hline
& & \textbf{En→En }& \textbf{En→En→L} & \textbf{En→L} & \textbf{L→L }\\
\hline
\multicolumn{2}{l}{\textbf{Initial solution}} & & & & \\
\hline
\multirow{2}{*}{\textbf{HRLs}} & {\tt Mistral-7b} & 99.8 & 99.8 & 99.8 & 99.8 \\
& Aya & 100.0 & 100.0 & 100.0 & 100.0 \\
\hline
\multirow{2}{*}{\textbf{LRLs}} & {\tt Mistral-7b} & 96.8 & 96.8 & \textbf{96.7} & 96.8 \\
& Aya & 98.6 & 98.6 & 98.6 & 98.6 \\
\hline
\multicolumn{2}{l}{\textbf{Revised solution}} & & & & \\
\hline
\multirow{2}{*}{\textbf{HRLs}} & {\tt Mistral-7b} & 93.1 & 93.7 & 94.9 & 94.3 \\
& Aya  & 92.0 & 92.9 & 93.4 & 93.0 \\
\hline
\multirow{2}{*}{\textbf{LRLs}} & {\tt Mistral-7b} & \textbf{90.6} & 92.1 & 92.3 & 91.8 \\
& Aya & 99.4 & 99.1 & 98.7 & 98.9 \\
\hline
\end{tabular}%
}
\caption{Mean language identification accuracy (in \%) of student models’ initial and revised solutions. Lowest value is bold for both initial and revised solutions.}
\vspace{-0.4cm} 
\label{tab:solution_comparison}
\end{table}

\paragraph{Impact of Multiple Hints on Student Gains (N>1).}
\label{para:hint-depth-analysis}
To evaluate whether providing multiple hints enhances student model performance, we consider single setup with L→L strategy—reflecting realistic multilingual tutoring scenarios—using {\tt LLaMA-3.3-70B} as the tutor and {\tt Mistral-7B} as the student. We extend the interaction up to N=5 iterations (i.e., up to five hints), terminating early if the student model produces the correct answer. Figure \ref{fig:5_hints} shows the relative improvement based on the number of hints required. For most languages, performance improves after the first hint, with Swahili showing the largest gain (over 70\%). In contrast, Telugu shows no improvement, suggesting that Mistral struggles significantly with this language—likely due to its English-centric training. Interestingly, the average number of hints needed across most languages is around two, indicating that a second hint often contributes meaningfully to student performance. Among high-resource languages, German requires the lowest number of hints, with the student typically improving after just one.

\section{Conclusion}
In this work, we present the first large-scale study involving 352 unique experiments on multilingual student–tutor interactions powered by LLMs, aimed at understanding the effect of language-specific hints across 11 typologically diverse languages, 4 models, and multiple prompting strategies. Our findings reveal that English-centric feedback can enhance student performance, but the most effective configurations vary depending on both the language and the LLM used. 
Additionally, we observed that even a few iterations of feedback can significantly improve problem solvability. This study offers key insights for designing equitable educational technologies and lays the groundwork for future research in multilingual feedback generation and evaluation. In future work, we aim to extend this research to subject areas beyond mathematics.

\section*{Limitations}

Although our work advances inclusive, multilingual LLM-based tutoring, several limitations remain and point to fruitful avenues for future research.

\paragraph{LLMs as Tutor--Student Simulators.}
As in prior studies \cite{macina2023mathdial, tran2025multi, wang-etal-2024-bridging}, we use LLMs to simulate both the tutor and the student, enabling large-scale, controlled experiments across languages and instructional settings. This strategy yields consistency and broad coverage, but it cannot capture the full diversity of real learners’ misconceptions, language proficiency, or problem-solving styles. Likewise, LLM tutors lack the nuanced pedagogical instincts of experienced tutors. Introducing human-in-the-loop evaluations, interactions with real students, and richer student models will increase validity and speed progress toward truly adaptive AI tutors.

\paragraph{Evaluating Hint Quality Beyond Student Gains.}
We treat a hint as “good’’ if it leads the student to the correct final answer -- that is, if it results in student gains. While practical and easy to measure, correctness alone misses key pedagogical dimensions such as conceptual scaffolding, clarity, and alignment with learning objectives. Expert reviews and rubric-based assessments by mathematics educators could supply these missing perspectives and help refine what counts as a high-quality hint.

\paragraph{Coarse Gain Measurement.}
Step-level verification of math reasoning with current LLMs is far harder than judging overall solution correctness \cite{daheim2024stepwiseverificationremediationstudent}. Consequently, we evaluate feedback with a binary metric -- does the student’s solution become fully correct or not? This ignores cases where feedback fixes the current error but a later, independent errors may still be present. Developing reliable, fine-grained metrics for partial progress is an important direction for future work.

\paragraph{Lack of Phase-wise Evaluation.}
Our pipeline follows the two-phase paradigm of first verifying the student’s work and then generating a hint \cite{macina2023mathdial}. Ideally, each phase should be evaluated separately. Yet automated assessment of both verification quality and hint usefulness is still unreliable; adding further sub-steps may boost overall performance but it compounds the evaluation challenge.

\paragraph{Limited Data and Language Coverage.}
Translating math-word problems is non-trivial: real-world contexts must be preserved, and many cultural references lack direct equivalents \cite{shi2022mgsm}. We therefore rely on the manually curated MGSM dataset, which contains only 250 problems per language across 11 languages. While sufficient for our experiments, this scale limits analyses such as comparing gains across typologically related languages or training models on parallel corpora to induce language-agnostic pedagogy. Expanding high-quality, multilingual datasets, especially for low-resource languages, remains a pressing need.

\section*{Acknowledgments}
We thank the supercomputing center at MBZUAI for providing resources for this research.

\bibliography{custom}

\newpage

\appendix

\section{Detailed Results}
\label{sec:det_result}

\begin{figure*}[htbp]
    \centering
    \begin{subfigure}[t]{0.48\linewidth}
        \centering
        \includegraphics[width=\linewidth]{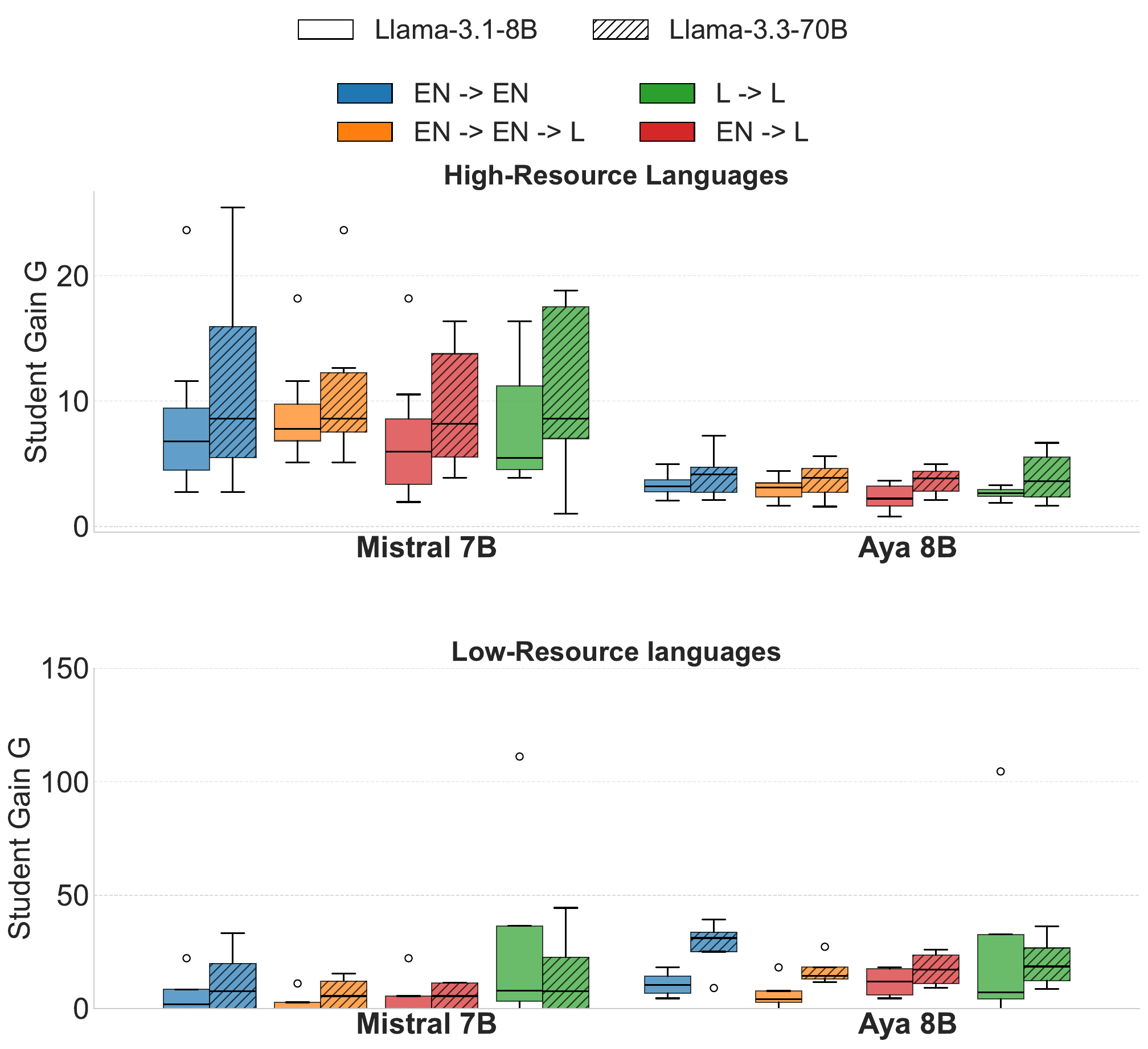}
        \caption{Multilingual prompting}
        \label{fig:native_student_prompt}
    \end{subfigure}
    \hfill
    \begin{subfigure}[t]{0.48\linewidth}
        \centering
        \includegraphics[width=\linewidth]{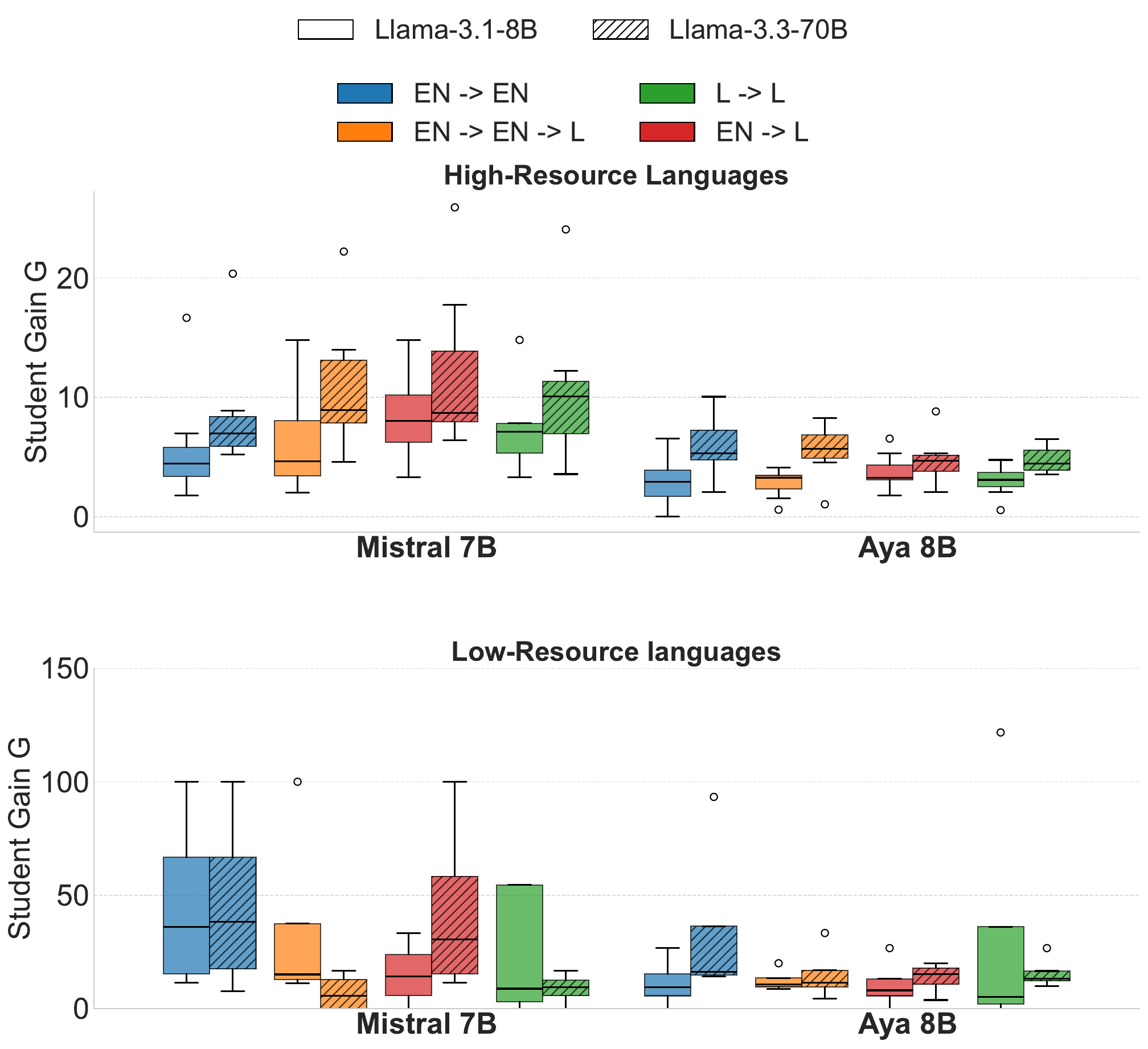}
        \caption{English-only prompting}
        \label{fig:english_student_prompt}
    \end{subfigure}
    \caption{Relative improvement (Student gain G) in multilingual student-tutor interaction.}
    \label{fig:box_plot_improvement}
\end{figure*}

Figure~\ref{fig:box_plot_improvement} provides a comprehensive summary of the results obtained from our extensive experimental setup.

\section{Dataset Representation}
\label{sec:app-data-representation}

The {MGSM} \cite{shi2022mgsm} dataset is built on top of the first 250 math problems from the {GSM8K} dataset \cite{Cobbe2021TrainingVT}. In order to make sure that this subset is representative of the larger set from {GSM8K}, we perform a feature level comparison between the two sets. For this, we borrow the feature set from \citet{srivatsa-kochmar-2024-makes} spanning the phrasing of the math problem in English, the count and nature of math operations and arguments involved in the gold solution, and the count of variables which require world knowledge. After generating the feature values for all English questions from {MGSM} and {GSM8K}, we compare their mean values -- see Table \ref{tab:ttest}. The low $t$-statistics and high p-values for corresponding pairwise $t$-tests indicate that there is not a significant difference between the two sets along any of the features.

\begin{table}[]
\centering
\resizebox{\columnwidth}{!}{%
\begin{tabular}{@{}lcccc@{}}
\toprule
\multirow{2}{*}{\textbf{Feature}} & \multicolumn{2}{l}{\textbf{Mean Feature Value}} & \multirow{2}{*}{\textbf{T-Statistic}} & \multirow{2}{*}{\textbf{p-value}} \\ \cmidrule(lr){2-3}
 & \textbf{{\tt GSM8K}} & \textbf{{\tt MGSM}} &  &  \\ \midrule
{\tt Gx\_mean\_numerical\_word\_rank} & 28638.85 & 28661.13 & -0.413 & 0.680 \\
{\tt Gx\_word\_arg\_count} & 0.64 & 0.68 & -0.407 & 0.684 \\
{\tt Gx\_world\_knowledge} & 1.07 & 1.06 & 0.140 & 0.889 \\
{\tt Qx\_constituency\_tree\_depth} & 10.97 & 10.96 & 0.044 & 0.965 \\
{\tt Qx\_flesch\_kinkaid\_grade} & 4.19 & 4.18 & 0.071 & 0.943 \\
{\tt Qx\_flesch\_reading\_ease} & 88.93 & 89.10 & -0.211 & 0.833 \\
{\tt Qx\_mean\_numerical\_word\_rank} & 22739.56 & 22592.54 & 0.664 & 0.507 \\
{\tt Qx\_mean\_word\_rank} & 10664.46 & 10596.28 & 0.466 & 0.641 \\
{\tt Qx\_multi\_np\_count} & 0.44 & 0.37 & 0.884 & 0.377 \\
{\tt Qx\_np\_count} & 18.52 & 18.62 & -0.208 & 0.835 \\
{\tt Qx\_prp\_count} & 1.82 & 1.90 & -0.688 & 0.492 \\
{\tt Qx\_sentence\_length} & 3.46 & 3.44 & 0.134 & 0.893 \\
{\tt Qx\_token\_length} & 67.43 & 67.49 & -0.034 & 0.973 \\
{\tt Qx\_unique\_np\_count} & 3.50 & 3.41 & 0.862 & 0.389 \\
{\tt Qx\_word\_arg\_count} & 1.11 & 1.17 & -0.599 & 0.549 \\
{\tt Qx\_word\_length} & 46.91 & 46.90 & 0.008 & 0.994 \\ \bottomrule
\end{tabular}%
}
\caption{Pairwise T-Test results between feature level mean values for {GSM8K} and {MGSM}.}
\label{tab:ttest}
\end{table}

\section{Comparative Analysis of Zero-Shot Performance Across Models}
\label{sec:compare_anlysis}
To select our student and tutor models, we initially evaluated five models -- {\tt Mistral-7B, Aya-8b, LLaMA-3.1-8B,LLaMA-3.3-70B}, and {\tt Deepseek-R1-LLaMA-distill-70B}\footnote{Via Together.ai: \url{https://www.together.ai/models/deepseek-r1-distilled-llama-70b-free}} -- using zero-shot prompting across both setups. We opted for standard zero-shot prompting over zero-shot Chain-of-Thought (CoT) prompting~\cite{Jin2024ZeroShotCR}, as CoT led to correct answers, whereas we sought student models that produce a balanced mix of correct and incorrect responses. Based on these criteria, we selected {\tt Aya-8b} and {\tt Mistral-7B} as student models, as shown in Table~\ref{tab:multilingual_performance}, which reports their balanced zero-shot accuracy—the baseline for subsequent comparisons. Table~\ref{tab:multilingual_performance} also shows that {\tt LLaMA-3.3-70B} achieves the highest accuracy across both setups and outperforms {\tt Deepseek-R1-LLaMA-distill-70B}, particularly with better output structure in low-resource languages. We therefore selected {\tt LLaMA-3.3-70B} and {\tt LLaMA-3.1-8B} as tutor models.

\begin{table*}[t]
\centering
\small
\setlength{\tabcolsep}{4pt}
\renewcommand{\arraystretch}{1.2}
\begin{tabular}{l|ccccc}
\hline
\multicolumn{6}{c}{\textbf{English only prompting: when student input prompt is in English language}} \\
\hline
\textbf{Languages} & \textbf{Llama 3.1-8B} & \textbf{Aya-8b} & \textbf{Mistral-7B} & \textbf{Llama-3.3-70B} & \textbf{Deepseek-R1-Llama-distill-70B} \\
\hline
English & 85.2 & 77.2 & 60.8 & 93.2 & 85.2 \\
Spanish & 76.8 & 72.0 & 46.0 & 86.0 & 80.8 \\
French & 74.8 & 68.0 & 44.8& 86.4 & 78.4 \\
German & 74.8 & 68.0 & 40.0 & 86.8 & 82.4 \\
Russian & 72.4 & 76.8 & 43.6 & 90.8 & 80.8 \\
Chinese & 72.4 & 67.2 & 36.0 & 88.0 & 84.0 \\
Japanese & 54.8 & 61.6 & 21.6 & 84.8 & 83.2 \\
Thai & 64.8 & 21.2 & 20.8 & 87.6 & 88.4 \\
Swahili & 60.8 & 9.2 & 3.6 & 84.8 & 79.2\\
Bengali & 60.0 & 28.0 & 9.6 & 85.6 & 79.2 \\
Telugu & 53.2 & 6.0 & 0.4 & 81.6 & 58.0 \\
\hline
\textbf{AVG} & 68.2\% & 50.5\% & 29.7\% & \textbf{86.9}\% & 80.0\% \\
\hline
\multicolumn{6}{c}{\textbf{Multilingual prompting: when student input prompt is in native language}} \\
\hline
\textbf{Languages} & \textbf{Llama 3.1-8B} & \textbf{Aya-8b} & \textbf{Mistral-7B} & \textbf{Llama-3.3-70B} & \textbf{Deepseek-R1-Llama-distill-70B} \\
\hline
English & 85.2 & 77.6 & 60.4& 93.6 & 82.8 \\
Spanish & 76.8 & 78.4 & 41.2 & 89.6 & 82.0 \\
French & 74.8 & 73.2 & 44.0 & 80.0 & 77.6 \\
German & 74.8 & 73.6 & 40.4 & 88.8 & 80.0 \\
Russian & 72.4 & 76.0 & 39.2 & 93.2 & 82.8 \\
Chinese & 72.4 & 72.0 & 38.0 & 88.0 & 86.8\\
Japanese & 54.8 & 64.4 & 22.0 & 84.8 & 83.2 \\
Thai & 64.8 & 18.4 & 10.4 & 86.8 & 83.2 \\
Swahili & 60.8 & 8.8 & 3.6 & 87.6 & 82.0 \\
Bengali & 60.0 & 20.4 & 9.2 & 85.6 & 81.6 \\
Telugu & 53.2 & 4.4 & 0.4 & 84.8 & 75.2 \\
\hline
\textbf{AVG} & 68.2 & 51.5 & 28.1 & \textbf{87.5} & 81.6\\
\hline
\end{tabular}
\caption{Zero-shot accuracy (\%) of language models across languages. These zero-shot scores serve as the baseline for each model. The table shows results for both English-only prompting (top) and multilingual prompting (bottom) setups. {\tt Llama-3.3-70B} consistently achieves the highest accuracy across most languages, demonstrating superior cross-lingual capabilities in zero-shot settings.}
\label{tab:multilingual_performance}
\end{table*}
\section{Implementation Details}
\label{sec:impl_details}
We set the temperature to 0 for the student models to ensure deterministic outputs, while the tutor models use a temperature of 1 to encourage diverse hint generation. GPT-4o \citep{Hurst2024GPT4oSC} was used with a temperature of 0 to evaluate the correctness of both the initial and revised answers by comparing them to the gold answer, as prior work has shown that temperatures above 0.2 can lead to unreliable results \citep{pmlr-v264-tonga25a}. For the tutor models, we employed {\tt LLaMA-3.3-70B} and {\tt LLaMA-3.1-8B}. For the student models, we selected a monolingual model, {\tt Mistral-7B}, and a multilingual model, {\tt Aya-8b}. All models and their corresponding reproduction links are presented in Table~\ref{tab:models}.

\begin{table}[t]
    \centering
    \small
    \renewcommand{\arraystretch}{1.3}
    \begin{tabular}{p{2.5cm}p{4.5cm}}
        \toprule
        \textbf{Model} & \textbf{Reproduction Link} \\
        \midrule
        {\tt Llama-3.3-70B} & \url{https://www.together.ai/models/llama-3-3-70b-free} \\
        {\tt Mistral-7B} & \url{https://huggingface.co/mistralai/Mistral-7B-Instruct-v0.3} \\
        {\tt Llama-3.1-8B} & \url{https://huggingface.co/meta-llama/Llama-3.1-8B-Instruct} \\
        {\tt Aya-8b} & \url{https://huggingface.co/CohereLabs/aya-expanse-8b} \\
         GPT-4o & gpt-4o-2024-08-06 \\
        \bottomrule
    \end{tabular}
    \caption{Models used in our experiments with their corresponding links for reproducibility.}
    \label{tab:models}
\end{table}
\section{Prompts}
In this section, we present the prompts used in our experiments.
\subsection{Student prompts}
\begin{figure}[h!]
    \centering
    \includegraphics[width=0.48\textwidth]{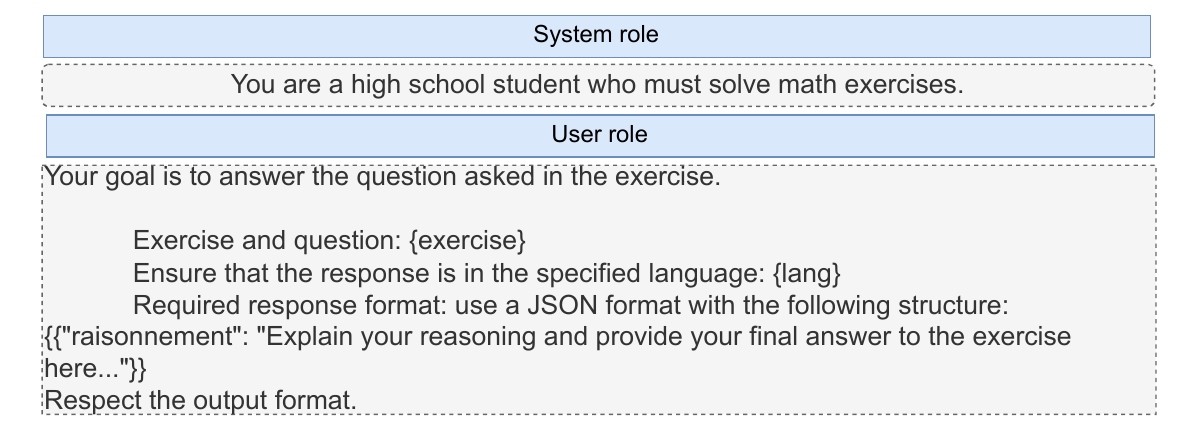}
    \caption{Prompt for generating a candidate solution.}
\label{fig:prompt_init_sol}
\end{figure}

\begin{figure}[h!]
    \centering
    \includegraphics[width=0.48\textwidth]{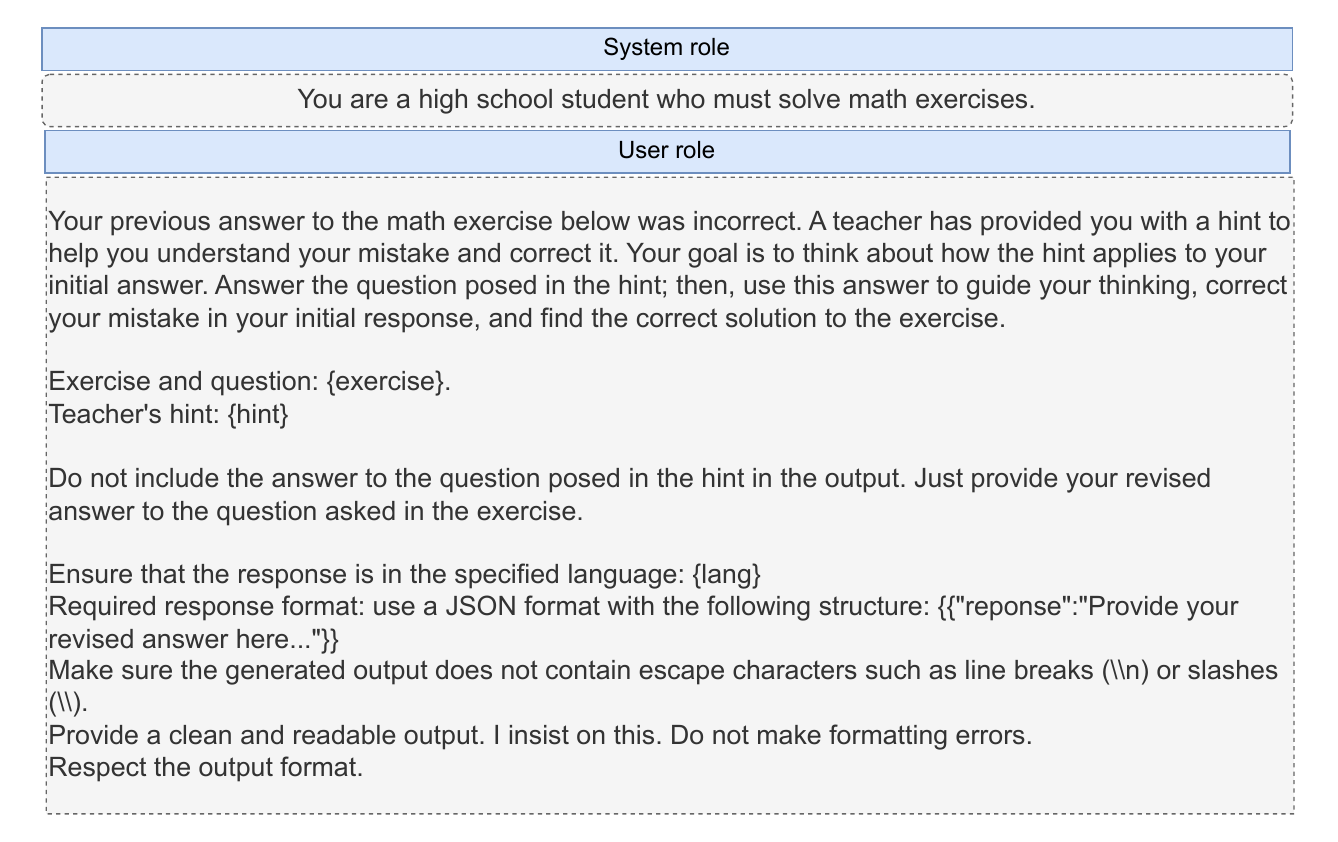}
\caption{Prompt for revising the initial candidate solution}
\label{fig:prompt_revise}
\end{figure}

Figure \ref{fig:prompt_init_sol} shows the base prompt for generating a candidate solution, while Figure \ref{fig:prompt_revise} displays the prompt for revising an initial answer via the student model. These prompts are used as-is in the English-only prompting setup. For the multilingual prompting setup, they are translated into the 10 target languages of the MGSM dataset.

\label{subsec:student_prompt}

\subsection{Hint generation prompt}
\label{subsec:teacher_prompt}
\begin{figure}[h!]
    \centering
    \includegraphics[width=0.48\textwidth]{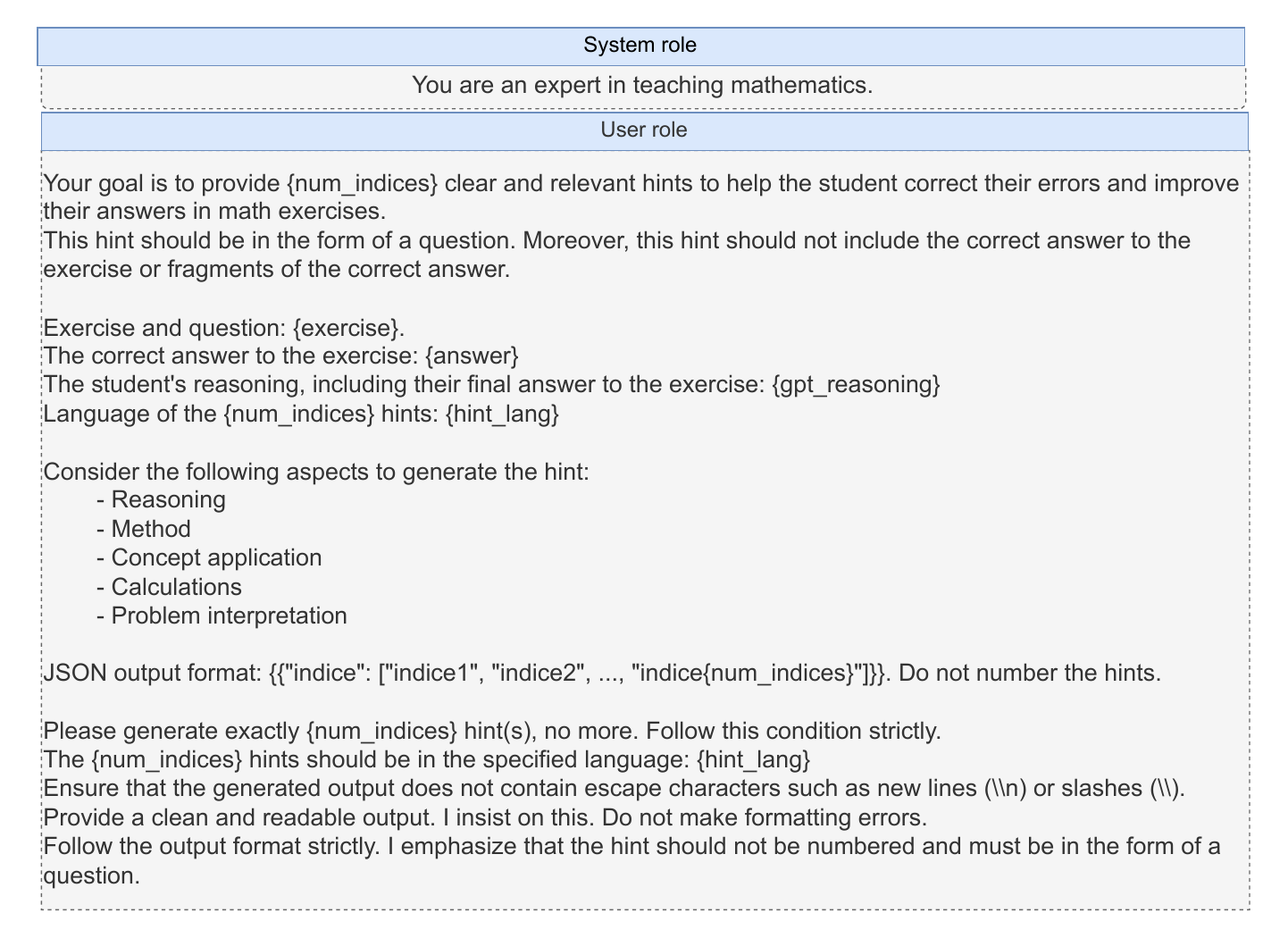}
\caption{Prompt used by the tutor model to generate hints.}
\label{fig:prompt_generate_hint}
\end{figure}

Figure \ref{fig:prompt_generate_hint} shows the base prompt used to generate a hint via the tutor model. For strategies where the tutor is prompted in English, the prompt is used as-is, with the variable \textbf{\textit{hint\_lang }} in the prompt replaced by either ``English'' (EN) or the exercise language (L), depending on the desired hint language. For strategy, where the tutor is prompted in the language of the exercise L, the prompt is translated into that language, and \textbf{\textit{hint\_lang}} is again set based on whether the hint should be in English or the exercise language.

\subsection{Prompts for Evaluating Student Outputs}
\label{subsec:eval_prompt}
\begin{figure}[h!]
    \centering
    \includegraphics[width=0.48\textwidth]{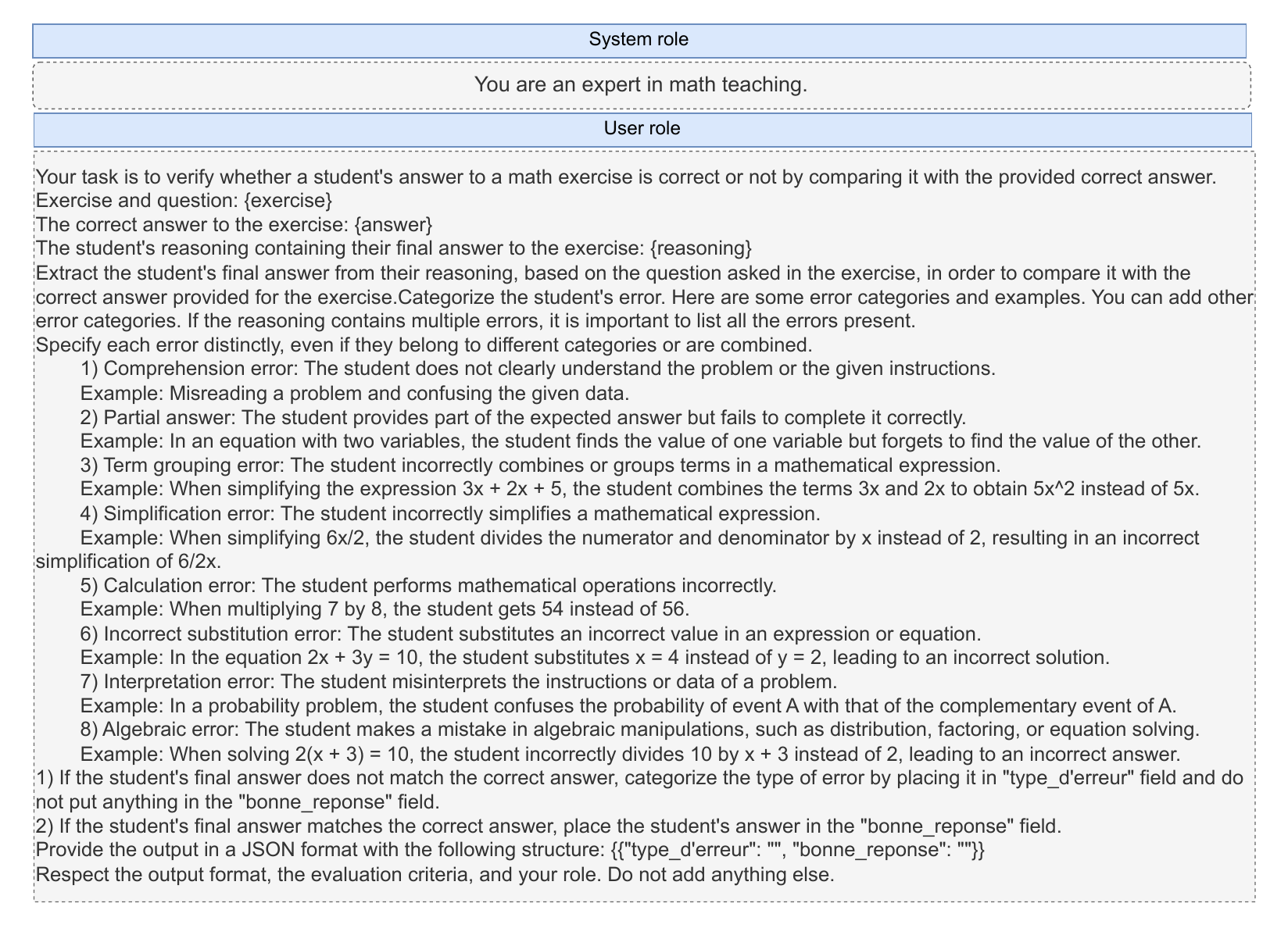}
\caption{Prompt employed by GPT-4o to assess candidate answer correctness.}
\label{fig:prompt_init_sol_eval}
\end{figure}

\begin{figure}[h!]
    \centering
    \includegraphics[width=0.48\textwidth]{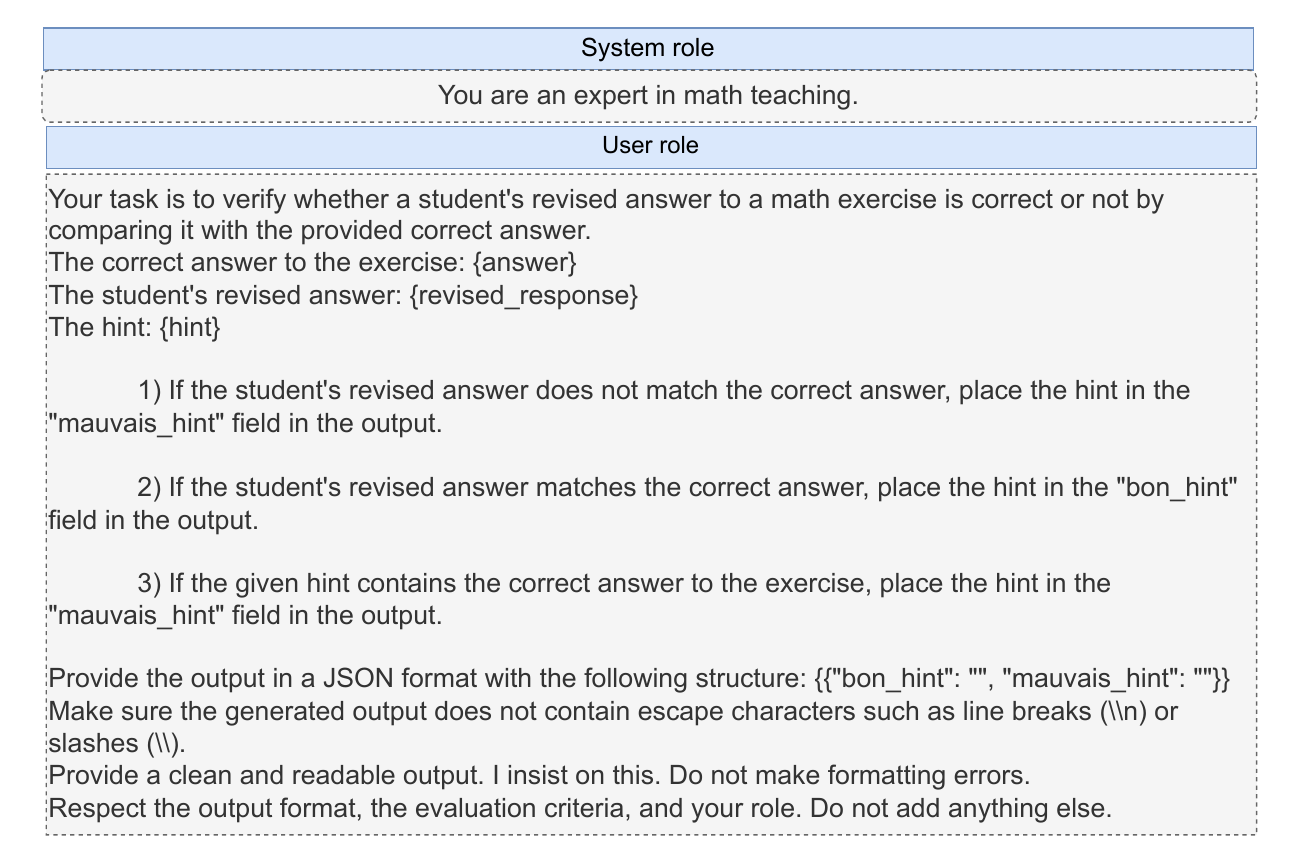}
\caption{Prompt employed by GPT-4o to assess the correctness of the revised solution and categorize hints.}
\label{fig:prompt_revised_eval}
\end{figure}

Figure \ref{fig:prompt_init_sol_eval} shows the prompt used to evaluate the initial candidate solution against the gold solution, while Figure \ref{fig:prompt_revised_eval} presents the prompt for evaluating the revised solution  via GPT-4o, after a hint is provided. We adopt the evaluation prompts from  \citet{pmlr-v264-tonga25a}, but omit their error type categorization for the initial candidate solution, as it is beyond the scope of our work. However, we retain their approach of classifying hints as good (if the revision is correct) or bad (if incorrect) to track hint effectiveness. 

\section{Analysis: Student gains across languages for the two different student models with {\tt LlaMA-3.3-70B}}
Table \ref{tab:student_gain_aya} and Table \ref{tab:student_gain_mistral} show the student gains of {\tt Aya-8b} and {\tt Mistral-7B}, respectively, after giving a hint, when we used {\tt LlaMA-3.3-70B} as the tutor model.

\begin{table}[t]
\centering
\setlength{\tabcolsep}{4pt}
\resizebox{\columnwidth}{!}{%
\begin{tabular}{llc}
\toprule
\textbf{Axis} & \textbf{Values} & \textbf{Count} \\
\midrule
\multirow{2}{*}{Languages} & en, bn, de, es, fr, ja,  & \multirow{2}{*}{11} \\
                           & ru, sw, te, th, zh & \\
\midrule
Student Prompts    & English-only, Multilingual              & 2 \\
Student Models     & {\tt Mistral-7B}, {\tt Aya-8B}          & 2 \\
\multirow{2}{*}{Hint Strategies} 
                  & EN$\rightarrow$EN, EN$\rightarrow$EN$\rightarrow$L, & \multirow{2}{*}{4} \\
                  & L$\rightarrow$L, EN$\rightarrow$L                & \\
tutor Models     & {\tt LLaMA-3.1-8B}, {\tt LLaMA-3.3-70B} & 2 \\
\midrule
\textbf{Total Configs} & 11 $\times$ 2 $\times$ 2 $\times$ 4 $\times$ 2 & \textbf{352} \\
\bottomrule
\end{tabular}
}
\caption{Overview of the experimental space.}
\label{tab:experiment-space}
\end{table}

\begin{figure}
    \centering
    \includegraphics[width=\linewidth]{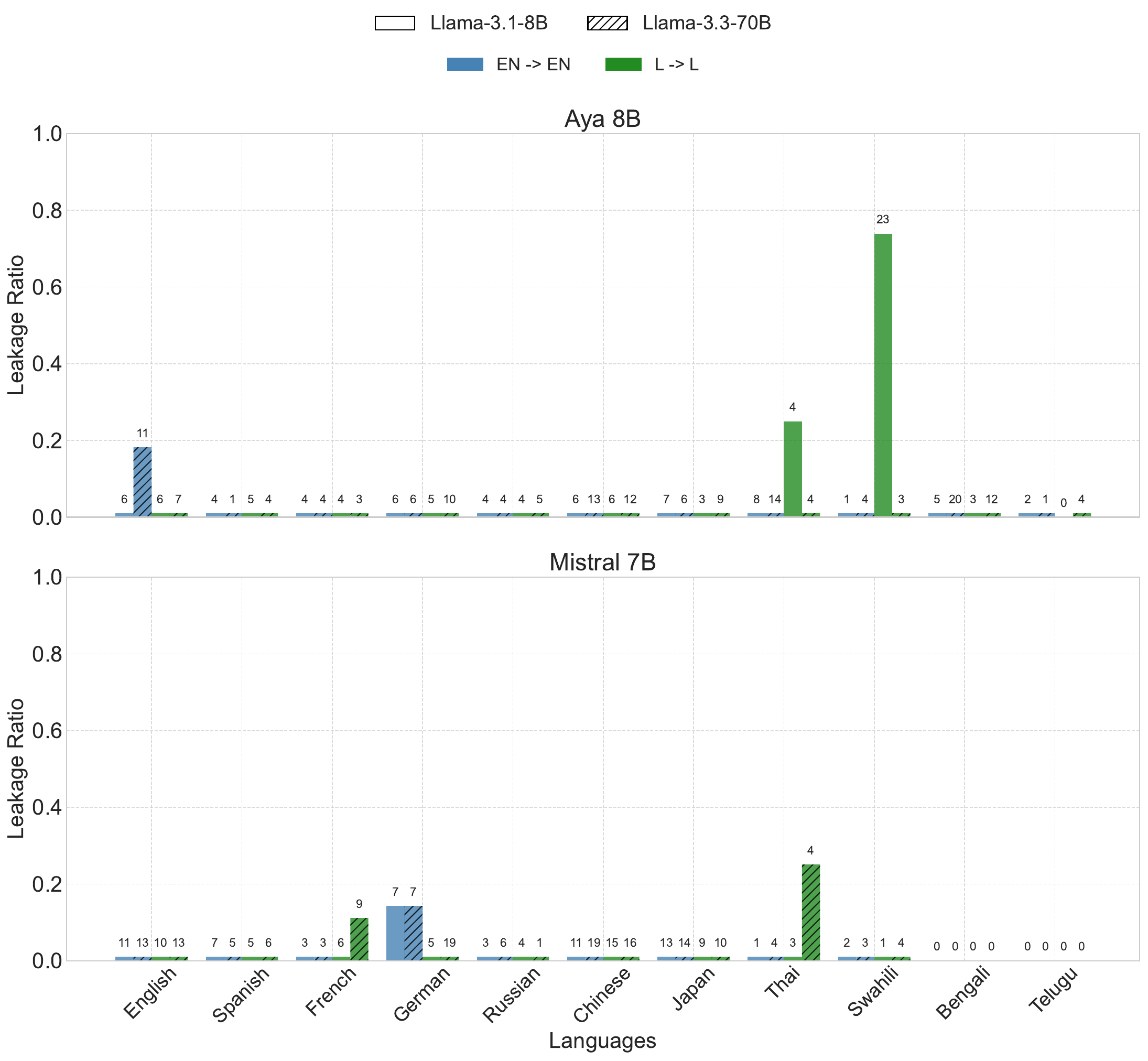}
   \caption{Leakage ratio per language; numbers above bars indicate total helpful hints.}
    \label{fig:leakage}
\end{figure}

\begin{table}[H]
\centering
\resizebox{0.8\columnwidth}{!}{%
\begin{tabular}{lc@{\hspace{2em}}lc}
\toprule
\textbf{Language} & \textbf{BLEU} & \textbf{Language} & \textbf{BLEU} \\
\midrule
Spanish & 62.7 & Chinese & 40.1 \\
French & 54.7 & Japanese & 45.6 \\
German & 46.7 & Thai & 40.0 \\
Russian & 49.3 & Swahili & 45.5 \\
 Bengali &  43.0 & Telugu & 39.5 \\
\bottomrule
\end{tabular}%
}
\caption{BLEU scores by language}
\label{tab:bleu_scores}
\end{table}

\begin{table*}[t]
\centering
\renewcommand{\arraystretch}{1.2}
\captionsetup{justification=centering}
\resizebox{\textwidth}{!}{%
\begin{tabular}{@{}c@{\hspace{1cm}}c@{}} 
\begin{subtable}[t]{0.47\textwidth}
\centering
\begin{tabular}{llcccc}
\toprule
\multicolumn{2}{c}{} & \multicolumn{4}{c}{\textbf{English-only prompting}} \\
\cmidrule(lr){3-6}
\multicolumn{2}{c}{} & EN$\rightarrow$EN & EN$\rightarrow$EN$\rightarrow$L & EN$\rightarrow$L & L$\rightarrow$L \\
\midrule
\multirow{7}{*}{\textbf{HRLs}} 
& English & 7.9 & 7.9 & 7.9 & 5.9 \\
& Spanish & 5.2 & 7.8 & 8.7 & 10.4 \\
& French & 5.4 & 8.9 & 8.0 & 3.6 \\
& German & 7.0 & 14.0 & 10.0 & 8.0 \\
& Russian & 6.4 & 4.6 & 6.4 & 10.1 \\
& Chinese & 8.9 & 12.2 & 17.8 & 12.2 \\
& Japanese & 20.4 & 22.2 & 25.9 & 24.1 \\
\midrule
\multirow{4}{*}{\textbf{LRLs}} 
& Thai & 7.7 & 11.5 & 11.5 & 7.7 \\
& Swahili & 55.6 & 0.0 & 44.4 & 11.1 \\
& Bengali & 20.8 & 16.7 & 16.7 & 16.7 \\
& Telugu & 100.0 & 0.0 & 100.0 & 0.0 \\
\bottomrule
\end{tabular}
\end{subtable}
&
\begin{subtable}[t]{0.85\textwidth}
\centering
\begin{tabular}{llcccc}
\toprule
\multicolumn{2}{c}{} & \multicolumn{4}{c}{\textbf{Multilingual prompting}} \\
\cmidrule(lr){3-6}
\multicolumn{2}{c}{} & EN$\rightarrow$EN & EN$\rightarrow$EN$\rightarrow$L & EN$\rightarrow$L & L$\rightarrow$L \\
\midrule
\multirow{7}{*}{\textbf{}} 
& English & 8.6 & 8.6 & 6.0 & 8.6 \\
& Spanish & 4.9 & 7.8 & 3.9 & 5.8 \\
& French & 2.7 & 7.3 & 8.2 & 8.2 \\
& German & 11.9 & 11.9 & 13.9 & 18.8 \\
& Russian & 6.1 & 5.1 & 5.1 & 1.0 \\
& Chinese & 20.0 & 12.6 & 13.7 & 16.8 \\
& Japanese & 25.5 & 23.6 & 16.4 & 18.2 \\
\midrule
\multirow{4}{*}{\textbf{}} 
& Thai & 15.4 & 15.4 & 11.5 & 15.4 \\
& Swahili & 33.3 & 11.1 & 11.1 & 44.4 \\
& Bengali & 0.0 & 0.0 & 0.0 & 0.0 \\
& Telugu & 0.0 & 0.0 & 0.0 & 0.0 \\
\bottomrule
\end{tabular}
\end{subtable}
\end{tabular}%
}
\caption{Student gains G (\%) of {\tt Mistral-7B} after receiving a hint from {\tt LlaMA-3.3-70B} across English-only and multilingual prompting setups.}
\label{tab:student_gain_mistral}
\end{table*}

\begin{table*}[t]
\centering
\renewcommand{\arraystretch}{1.2}
\captionsetup{justification=centering}
\resizebox{\textwidth}{!}{%
\begin{tabular}{@{}c@{\hspace{1cm}}c@{}}
%
\begin{subtable}[t]{0.47\textwidth}
\centering
\begin{tabular}{llcccc}
\toprule
\multicolumn{2}{c}{} & \multicolumn{4}{c}{\textbf{English-only prompting}} \\
\cmidrule(lr){3-6}
\multicolumn{2}{c}{} & EN$\rightarrow$EN & EN$\rightarrow$EN$\rightarrow$L & EN$\rightarrow$L & L$\rightarrow$L \\
\midrule
\multirow{7}{*}{\textbf{HRLs}} 
& English & 5.7 & 5.7 & 3.1 & 4.1 \\
& Spanish & 5.0 & 7.2 & 5.0 & 4.4 \\
& French & 8.8 & 6.5 & 8.8 & 5.9 \\
& German & 5.3 & 5.3 & 4.7 & 5.3 \\
& Russian & 2.1 & 1.0 & 2.1 & 3.6 \\
& Chinese & 10.1 & 8.3 & 5.3 & 3.6 \\
& Japanese & 4.5 & 4.5 & 4.5 & 6.5 \\
\midrule
\multirow{4}{*}{\textbf{LRLs}} 
& Thai & 15.1 & 11.3 & 3.8 & 13.2 \\
& Swahili & 17.4 & 4.3 & 13.0 & 13.0 \\
& Bengali & 14.3 & 11.4 & 17.1 & 10.0 \\
& Telugu & 93.3 & 33.3 & 20.0 & 26.7 \\
\bottomrule
\end{tabular}
\caption{English-only prompting}
\end{subtable}
&
%
\begin{subtable}[t]{0.85\textwidth}
\centering
\begin{tabular}{llcccc}
\toprule
\multicolumn{2}{c}{} & \multicolumn{4}{c}{\textbf{Multilingual prompting}} \\
\cmidrule(lr){3-6}
\multicolumn{2}{c}{} & EN$\rightarrow$EN & EN$\rightarrow$EN$\rightarrow$L & EN$\rightarrow$L & L$\rightarrow$L \\
\midrule
\multirow{7}{*}{\textbf{HRLs}} 
& English & 4.1 & 4.1 & 3.1 & 3.6 \\
& Spanish & 5.1 & 5.1 & 2.6 & 2.0 \\
& French & 2.2 & 2.7 & 3.8 & 1.6 \\
& German & 3.3 & 2.7 & 4.9 & 5.4 \\
& Russian & 2.1 & 1.6 & 2.1 & 2.6 \\
& Chinese & 7.2 & 3.9 & 3.9 & 6.7 \\
& Japanese & 4.3 & 5.6 & 5.0 & 5.6 \\
\midrule
\multirow{4}{*}{\textbf{LRLs}} 
& Thai & 30.4 & 15.2 & 26.1 & 8.7 \\
& Swahili & 31.8 & 13.6 & 22.7 & 13.6 \\
& Bengali & 39.2 & 11.8 & 11.8 & 23.5 \\
& Telugu & 9.1 & 27.3 & 9.1 & 36.4 \\
\bottomrule
\end{tabular}
\caption{Multilingual prompting}
\end{subtable}
\end{tabular}%
}
\caption{Student gain G (\%) of {\tt Aya-8b} after receiving a hint from {\tt LlaMA-3.3-70B} in the two setups.}
\label{tab:student_gain_aya}
\end{table*}

\end{document}